\documentclass[sigconf]{acmart}

\AtBeginDocument{%
  }

\usepackage{balance}

\setcopyright{acmcopyright}
\copyrightyear{2023}
\acmYear{2023}
\acmDOI{XXXXXXX.XXXXXXX}

\acmConference[ICAIF '23]{Make sure to enter the correct
  conference title from your rights confirmation emai}{November 27--29,
  2023}{New York, NY}

\acmPrice{15.00}
\acmISBN{978-1-4503-XXXX-X/18/06}

\usepackage{xcolor}
\definecolor{c1}{rgb}{1.0, 0.89, 0.77}
\definecolor{g1}{rgb}{0.67, 0.88, 0.69}
\definecolor{r1}{rgb}{1.0, 0.72, 0.77}
\definecolor{cgreen}{rgb}{0.0, 0.42, 0.24}
\definecolor{dmagenta}{rgb}{0.55, 0.0, 0.55}

\usepackage{colortbl}
\usepackage{array}
\usepackage{fancyhdr}
\newcolumntype{T}[1]{%
    >{\centering\arraybackslash\hspace{0pt}}p{#1}}%
\newcolumntype{L}[1]{>{\raggedright\let\newline\\\arraybackslash\hspace{0pt}}m{#1}}

\newcommand{\bx}{\mathbf{x}}
\newcommand{\sx}{\boldsymbol{v}}

\usepackage{amsmath}
\usepackage{graphicx}
\usepackage{subcaption}

\fancypagestyle{firststyle} 
{
  \fancyhf{}
  
  \fancyfoot[C]{\Large\textcolor{blue}{\textbf{Best Industry Paper Award at the International Conference on AI in Finance (ICAIF) 2023}}}
}

\begin{document}

\title{Multi-Modal Financial Time-Series Retrieval Through Latent Space Projections}

\author{Tom Bamford}
\authornote{All authors contributed equally to this research.}
\affiliation{%
  \institution{J.P. Morgan AI Research}
  \country{ }
}

\author{Andrea Coletta}
\authornotemark[1]
\affiliation{%
  \institution{J.P. Morgan AI Research}
  \country{ }
}
\author{Elizabeth Fons}
\authornotemark[1]
\affiliation{%
  \institution{J.P. Morgan AI Research}
  \country{ }
}
\author{Sriram Gopalakrishnan}
\authornotemark[1]
\affiliation{%
  \institution{J.P. Morgan AI Research}
  \country{ }
}
\author{Svitlana Vyetrenko}
\affiliation{%
  \institution{J.P. Morgan AI Research}
  \country{ }
}
\author{Tucker Balch}
\affiliation{%
  \institution{J.P. Morgan AI Research}
  \country{ }
}
\author{Manuela Veloso}
\affiliation{%
  \institution{J.P. Morgan AI Research}
  \country{ }
}

\newcommand{\dored}[1]{\textcolor{red}{#1}}

\begin{abstract}
Financial firms commonly process and store billions of time-series data, generated continuously and at a high frequency. To support efficient data storage and retrieval, specialized time-series databases and systems have emerged. These databases support indexing and querying of time-series by a constrained Structured Query Language(SQL)-like format to enable queries like \textit{"Stocks with monthly price returns greater than 5\%"}, and expressed in rigid formats. However, such queries do not capture the intrinsic complexity of high dimensional time-series data, which can often be better described by images or language (e.g., \textit{"A stock in low volatility regime"}). Moreover, the required storage, computational time, and retrieval complexity to search in the time-series space are often non-trivial. 
In this paper, we propose and demonstrate a framework to store multi-modal data for financial time-series in a lower-dimensional latent space using deep encoders, such that the latent space projections capture not only the time series trends but also other desirable information or properties of the financial time-series data (such as price volatility). Moreover, our approach allows user-friendly query interfaces, enabling natural language text or sketches of time-series, for which we have developed intuitive interfaces.
We demonstrate the advantages of our method in terms of computational efficiency and accuracy on real historical data as well as synthetic data, and highlight the utility of latent-space projections in the storage and retrieval of financial time-series data with intuitive query modalities.
\end{abstract}

\maketitle

\thispagestyle{firststyle}

\section{Introduction}

The increasing usage of data-hungry applications, including AI/ML algorithms, has brought significant changes to storage systems required to accommodate the growing volume, velocity, variety of data and queries. 
In particular, financial firms are required to process and store billions of time-series (TS). These TS are stored to satisfy compliance requirements, to provide business and clients with historical data, and to support data-driven algorithms. 
Such TS are frequently accessed with varying requirements, and correspondingly, queries. To support efficient data storage and retrieval, specialized TS databases and systems have emerged~\cite{taylor_2022,winston2022time,timescale,influxdb,hao2021ts,shah2022performance}.
While these databases support efficient indexing and querying of TS, they do not directly target two important aspects of financial TS: the intrinsic complexity of high-frequency financial data; and the required retrieval modalities and difficulty for the end users.
In fact, such databases commonly offer queries based on some fixed set of  properties and using SQL style queries; for example, "Stocks with daily price returns $\leq$ 5\%". However, statistical properties of real price series can be non-trivial and challenging for users to express in such rigid formats~\cite{bouchaud2018trades}. Therefore, financial users would require a fast and easy range of query modalities that facilitate the input of financial TS features, called stylized facts (e.g., volatility or correlation)~\cite{vyetrenko2020get}. 

%
Some academic work proposes methods to: analyze and retrieve massive heterogeneous datasets; manage TS data streams from Internet of Things (IoT) devices; or find uncommon and interesting patterns w.r.t. historical database~\cite{yang2019edgedb,bitincka2010optimizing,keogh2002finding}. However, to the best of our knowledge, these works focus on their specific applications or problems. They do not target or apply to financial TS, which have their own specific statistical properties and requirements~\cite{bouchaud2018trades}.  

\begin{figure*}[t]
  \centering
  \begin{subfigure}[b]{0.45\textwidth}
    \centering
    \includegraphics[width=\textwidth]{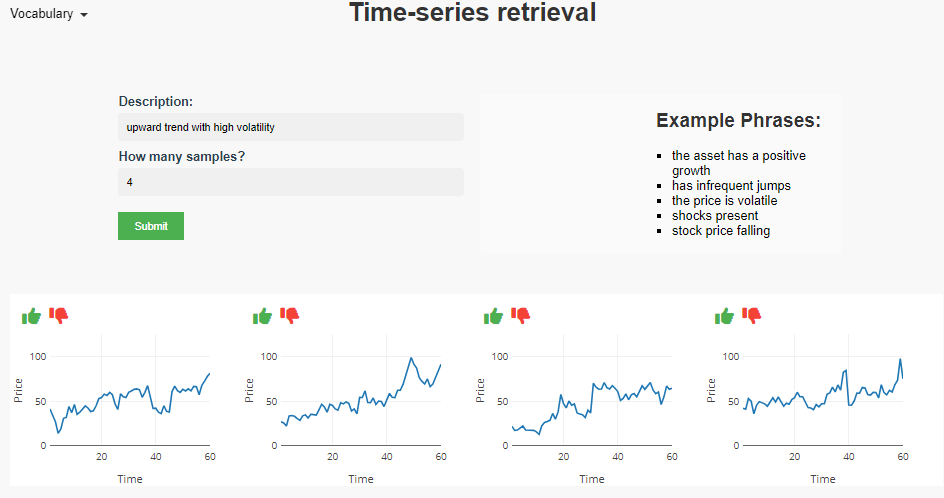}
    \caption{Prototype interface for text-based TS retrieval; allows user to describe the TS properties when searching historical data }
    \label{subfig:image1}
  \end{subfigure}
  \hfill
  \begin{subfigure}[b]{0.45\textwidth}
    \centering
    \includegraphics[width=\textwidth]{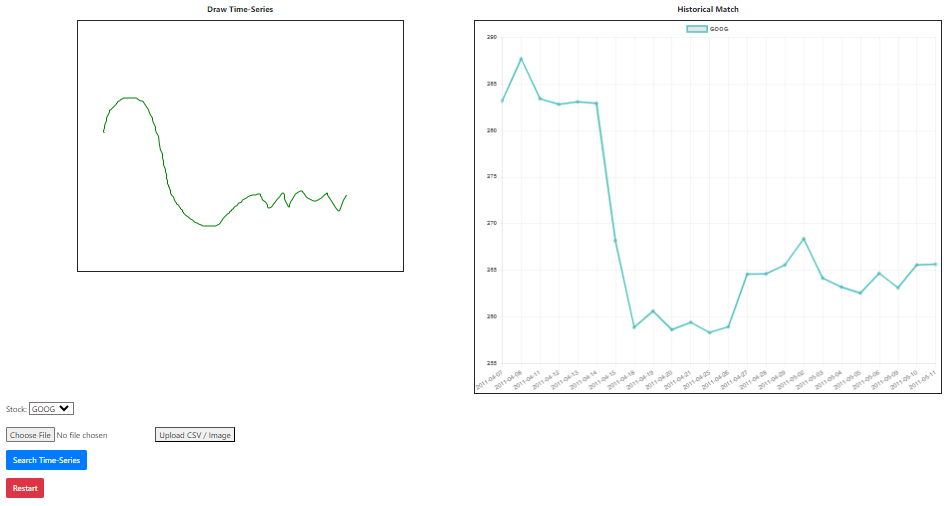}
    \caption{Prototype interface for sketch-based TS retrieval; allows user to draw a sketch of a hypothetical TS to search historical data}
    \label{subfig:image2}
  \end{subfigure}
  \caption{Interfaces developed for natural language and sketch based queries.}
  \label{fig:main_figure}
\end{figure*}

Motivated by the lack of dedicated solutions for the financial domain, and by leveraging the recent advancements in machine-learning techniques, we propose a framework for efficient and easy multi-modal TS retrieval using latent space projections for financial TS. 
In particular, our main contribution is a framework for using Deep Encoder Networks for storing and retrieving financial TS data across different modalities, and retaining finance-pertinent information in the latent space. Our approach enables one to learn a shared embedding space between multi-modal data (e.g., text to TS), and allow retrieval to be carried out using different query-modes (natural language, TS sketches, images, etc). We use such a learned embedding space to index the historical TS data (i.e., we create a database of <encoding, historical data>). The database can then be used to retrieve TS using the different input modalities, and we evaluate our methods using real historical data and synthetic data. We demonstrate our framework in two instances of the general approach of using deep encoders; one for queries by natural language (\textbf{text-based retrieval}), and one for queries by sketches of TS (\textbf{sketch-based retrieval}). Interfaces for these modalities are shown in Figure \ref{fig:main_figure}. The user can search historical TS by textual queries: \textit{"A stock with high volatility and increasing price"}(Figure~ \ref{subfig:image1}, and our encoder is trained to pay attention to pertinent-financial descriptors like ``high volatility''. Alternatively, one can draw the general sketch of the TS as a search input as it is often the easiest to visualize a trend of the TS of interest (Figure~ \ref{subfig:image2} for example).

Our text-based retrieval method leverages a pre-trained language encoder and a pre-trained visual encoder to combine visual and textual features, whose latent space are aligned using cosine similarity in the fine-tuning phase. The proposed system is derived from the OpenAI CLIP network \cite{Radford_CLIP} which has shown state-of-the-art performance in content based image retrieval. In order to incorporate sketch-based functionality, our second model uses autoencoders to encode a TS sketch and the volatility trend seen in the sketch into a latent-space, so subsequent queries can be matched with more than just the trend information. This is not limited to just volatility, but extensible to different properties of price TS data.

We experimentally evaluate our approach and show its benefits in terms of computational time and accuracy. We compare it against multiple baselines including standard TS retrieval approaches, and latent-space retrieval using other dimensionality reduction methods, specifically Uniform Manifold Approximation and Projection (UMAP)~\cite{mcinnes2018umap}. We show that our approach can reduce the retrieval time, and provide competitive if not better accuracy than the baselines while supporting user-friendly modalities for querying. Finally, it is worth mentioning the flexibility of our approach, which can support additional modalities.

In summary the contributions of our work are as follows:
\\
$\bullet$ We introduce a deep encoding framework for multi-modal retrieval that is focused on financial TS. Our framework is adapted to the financial domain by encoding stylized facts of financial TS explicitly.
\\
$\bullet$ We demonstrate the functionality of our framework for text-based and sketch-based retrieval modalities
\\
$\bullet$ We evaluate and show the benefits of our work in terms of computational time and accuracy by comparing them against existing baselines and approaches.

\section{Related Work}
In this section we briefly survey existing, academic and commercial, work on TS retrieval.

\paragraph{\textbf{Commercial and open-source TS databases.}}
Among the most famous TS oriented databases we find InfluxDB~\cite{influxdb}, a popular open-source database with high-performance storage and retrieval for TS, providing continuous SQL-like queries and additional specific TS features like downsampling. TimescaleDB~\cite{timescale} is an open-source database built on top of PostgreSQL to have its reliability and robustness while  providing new storage optimization for TS. One of the most famous commercial TS database is KDB+~\cite{taylor_2022}, which supports high-frequency data analytics by organizing data into a hierarchical memory, which enables state-of-art performance and lower hardware cost~\cite{barez2023benchmarking}. Finally, Amazon recently introduced Timestream~\cite{winston2022time}, a cloud serverless TS database service to optimize in-memory and historical data, while also providing TS analytical tools including smoothing, approximation, and interpolation for users. 
Most of these databases focus on efficient indexing and storage of TS data. They store most older TS in more resilient and high capacity memories, while they keep only more recent and partial data in-memory. However, in contrast to us, they do not focus on providing multi-modal queries to ease the end users in retrieving complex financial TS data. In fact, existing databases often only support complex and slow queries, while our work aims at providing intuitive modalities for financial TS retrieval like natural language input and trend sketches. 

\paragraph{\textbf{Research Literature}}
Existing work on multi-modal search considers how to incorporate different modalities like text, and images to query databases~\cite{multimode_1_tautkute2019deepstyle}~\cite{multimode2_wang2016effective}. However, to the best of our knowledge, our work is the first to target multi-modal TS retrieval specifically for financial TS data. The general approach we take in our methods -- which is deep encoders with modifications to encode pertinent information for financial TS-- allow one to query with natural language text or sketches of TS. Existing literature work mostly focus only on specific problems (e.g., multi-variate binary encoding and retrieval \cite{zhu2020deep,song2018deep}, or tree index structure~\cite{assent2008ts})  or target specific domain-related issues~\cite{al2019time,liu2019high}. L. Bitincka et al.~\cite{bitincka2010optimizing} propose a database that can be used to index, search and analyze massive heterogeneous datasets; Y. Yang et al.~\cite{yang2019edgedb} focus on TS data streams from Internet of Things (IoT) devices and their management on edge nodes; and E. Keogh et al.~\cite{keogh2002finding} discuss the analysis of TS to find uncommon or interesting patterns w.r.t. historical database. However, to the best of our knowledge, previous work does not target or apply to multi-modal financial TS retrieval, which has specific requirements such as incorporating pertinent statistical properties of financial data~\cite{bouchaud2018trades} such as price volatility. 

For \textbf{text-based retrieval} of TS, very little prior work has been done in this area. However, a couple of relevant attempts to tackle this problem have been made~\cite{Agrawal_Querying_Shapes_Histories, Imani_Text_TS_Retrieval}. In both cases a limited, pre-defined vocabulary is introduced which is associated with specific numeric features of the TS. Such a pre-defined dictionary can then be used to convert textual queries into numeric searches to retrieve relevant TS. The limited nature of the vocabulary used, alongside the clunky nature and relative unintuitveness render these methods distinctly lacking when compared to our deep encoding approach in the scope of queries encompassed, particularly with regards to out-of-sample or ad-hoc querying. In addition, the approaches put forward were domain-agnostic; our approach can be applied both in a domain-agnostic fashion and tailored to financial TS specifically.

With respect to \textbf{sketching-based retrieval} for TS, there is prior work that targeted this problem~\cite{sketchTS_retrieval}~\cite{sketch2_mannino2018expressive}. However, in these methods, the approaches only match trends and artefacts (shapes) and do not consider any additional properties such as volatility of the TS (which we do). Our baseline methods cover the methods used in the aforementioned work and we modify the baselines to try and incorporate volatility into those methods in the experiments. 

\section{Problem Description}
We first define the TS space as $ \chi = \mathbb{R}^{L}$ where $L$ is the length of each TS. Our problem focuses on the retrieval of TS from a large input dataset $\mathbb{D} = \{\mathbf{x}^i\}_{i=1}^N$ of $N$ time series ($\bx^i\in\chi, i\in[1..N]$), using different modalities. Formally, our goal is to learn a function $f(q \in \chi,\mathbb{D},k) \rightarrow R \subset D $ that, given an input query $q$, retrieves the best $k$ matching TS from $\mathbb{D}$. For example, a text based input \textit{"A stock with high volatility and increasing price"} should retrieve a TS with such described statistical properties.  
In particular, financial TS have specific statistical properties; price series data (which we focus on) in particular has non-trivial properties that emerge from the mechanisms of individual actions and interactions in market micro-structure. Such statistical properties are often referred to as stylised facts~\cite{vyetrenko2020get,bouchaud2018trades}. These can be challenging to express with existing query frameworks for TS data. Our goal is to support efficient and intuitive query modalities to enable users to retrieve TS that match not just the trend but also financially-relevant properties, such as price volatility. Most importantly, we tackle the problem of multi-modal retrieval to empower users to retrieve TS using different types of input (e.g. text, sketches, images, and more). 

In the next section, we describe the general approach we take to enable multi-modal retrieval. For simplicity, we will consider only univariate TS, although the approach can be generalized to multi-variate TS. Likewise, we restrict the retrieval modalities to text-based search using natural language and TS sketch-based search, although the same approach we employ here can be applied to other modalities like images. 

\section{Dataset Construction}\label{database_construction}

\subsection{Synthetic stock price time series dataset}

There are no existing datasets that have financial TS paired with corresponding textual descriptions. This type of dataset is crucial to train a text-based retrieval model that can effectively extract meaningful information and establish a connection between these two data types. Therefore, we could generate synthetic TS and their corresponding textual descriptions by simulating stock prices~\cite{bouchaud2018trades,coletta2022learning,byrd2020abides}, for example using deep generative models~\cite{coletta2023constrained,yoon2019time}.  
Specifically, to generate synthetic stock data we utilize the discrete mean-reverting TS  which is frequently used to model financial markets \cite{wah_wellman} as well as biological processes \cite{biological} and is described by the equations below:
\begin{align*}
r_t = \max\{0, \kappa\bar{r}+ \left( 1 - \kappa \right) r_{t-1} + u_t\}, r_0 = \bar{r},
\end{align*}
where $\bar{r}$ is a mean value of the TS, $\kappa$ is a mean-reversion parameter and $u_t \sim \mathcal{N} (0, \sigma^2)$ is random noise added to the TS at each time step $t$. 
We bring stock directionality and a possibility of a large shock occurrence to the above generating process by introducing the concepts of trend and megashocks. Trend $T$ is added to $r_t$ at each time step $t$ to indicate the incline or decline of the stock value. As in \cite{byrd2019explaining}, megashocks are intended to represent the exogenous events that occur infrequently and can have significant impact on the generating process. Mathematically, megashocks can arrive at any time $t$ with probability of occurrence $p$, and are drawn from $\mathcal{N} (0, \sigma_{shock}^2)$ where $\sigma_{shock}>>\sigma$. Once the time series are generated, they are converted into 224x224 images by plotting on a pre-defined domain. We test three different approaches for auto-generation of captions.

\paragraph{Unfiltered}{
We associate the numerical value of each parameter $\bar{r}$, $\kappa$, $\sigma$, $T$, $p$, $\sigma_{shock}$ with a sentiment that describes it. For instance, close to zero values of trend $T$ can be described as "neutral, horizontal, non-increasing, flat, stable, unchanged"; larger positive values of $T$ can be described as "upward, growing, positive, increasing, rising, climbing, advancing"; smaller negative values of $T$ can be described as "declining, falling, sliding, sinking, plummeting, downward". Similarly, for high values of $\sigma$, the stock price TS can be described as "has strong variability", "has significant variations", "has
aggressive variations", "is unstable", "has high fluctuation", "is
noisy", "is variable"; whereas, for the low values of $\sigma$, the generated TS can be labeled as "has small volatility", "the stock shows a slight
variability", "the stock has negligible volatility", "has low
volatility", "the price remains stable". For a given image, we sample from a list of 3-5 semantically similar phrases to generate a specific phrase for each corresponding ground truth regime; each chosen feature phrase is then concatenated to give the final ground truth caption.}

\paragraph{Filtered}{
Due to the stochasticity and inter-dependence between parameters in our synthetic model, auto-generation of captions based on parameter value in some instances leads to surprising captioning results relative to expectations based on a simple `eye-test'. We found this particularly prominent in higher volatility settings. In order to ensure consistency in captioning, we carry out a post-process filtering step, in which generated TS are re-labelled in trend and shock probability regimes dependent on mathematically pre-defined conditions. For the trend, we fit a linear curve to the series and evaluate the appropriate regime based on the fitted gradient with respect to a threshold value. For shock regime filtering, we evaluate the gradient of the generated TS at each point and assign the shocked regime based on the presence of any gradient values above a threshold magnitude.

For the historic dataset we extend the filtering process to include also a volatility post-processing check. In this case, the volatility regime is assigned based on the average deviation with respect to the running TS value. This was found to give more consistent captioning than the traditional volatility definition in finance which uses the standard deviation of returns.
}

\paragraph{Filtered+}{
Given that the number of hand-crafted text descriptions for each configuration of parameters is limited, we augmented the text descriptions using ChatGPT. We did this by feeding ChatGPT a few pre-defined sentences for each feature regime, and asking it come up with a large number of semantically similar phrases. These were than saved to text files from which alternative captions could be drawn. In total this augmented the dataset by around 60-80 phrases for each feature regime, increasing the total number of phrases from 36 to over 500. This enables us to correspondingly increase the size of the TS dataset and reduce the possibility of overfitting. Due to the synthetic nature of the dataset creation, we are able to easily scale up to larger dataset sizes, going from 4000 to 16000 samples for training.
}

It is well known that low liquidity stocks have high volatility, while high liquidity stocks have low volatility \cite{bouchaud2018trades}.
Therefore, we can expand the captioning of the dataset generated above to differentiate between "high liquidity" and "low liquidity" stocks based on their volatility profiles. Note finally that the format of the caption attached to each image is a comma-separated statement about each feature of interest.

\subsection{Historical price time series dataset}

For our \textbf{text-based retrieval} method, we collected historical stock data from Yahoo Finance using their open-source Python package. The stock tickers selected were GOOG, AMD, INBX and DAWN for high volatility, medium volatility and low volatility (latter two) regimes respectively. We constructed a dataset of TS with 60 timesteps, using overlapped sampling with a specified window size. For the smaller dataset, we use only GOOG, AMD and INBX stocks to give 1500 TS in total, whilst for the larger dataset we include the additional DAWN data for the low volatility setting and modify the window size to double the number of TS to 3000. As with the synthetic dataset, assigned captions are comma-separated statements about each feature of interest, and TS are converted to 224x224 images through plotting using a fixed format.

For our \textbf{sketch-based retrieval} method, we similarly collected stock data for GOOG, AMD, and INBX with each trace of being of length 30 (corresponds to 1 month) and successive traces are obtained in increments of 5 time-steps. The dataset had 1516 traces which we found sufficient to show the relative performance costs between the baseline methods and using an autoencoder (AE).

\section{Methodologies}
\begin{figure*}[h!]
    \centering
    \includegraphics[width=0.9\linewidth]{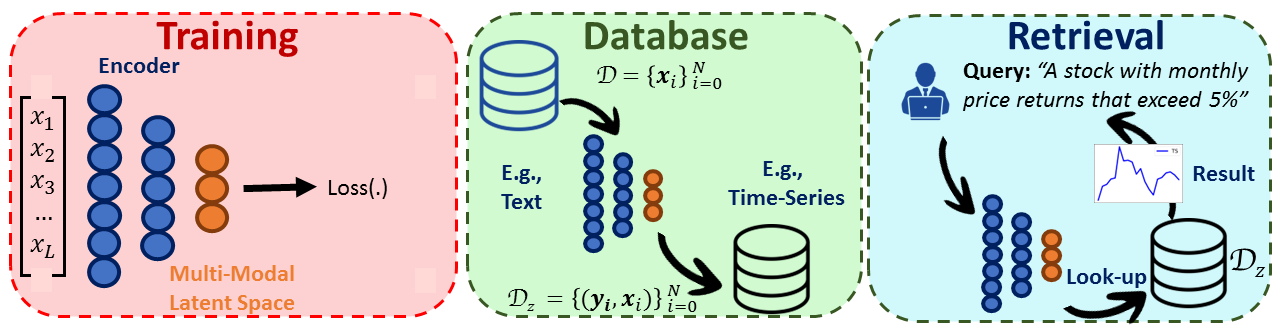}
    \caption{\small General model architecture for both retrieval methods. The Deep Encoders are trained with different modes of data against the loss to create a shared latent-space. A database containing the TS and the latent-space indexing is created. Finally, the user query is converted through the trained encoder into the latent space to query the database, and matching TS are retrieved.}
    \label{fig:model_arch}
\end{figure*}

In this section we introduce our approach for text-based retrieval and sketch-based retrieval of TS. Figure~\ref{fig:model_arch} shows the overall proposed framework. In particular, we propose the use of Deep Encoder Networks~\cite{goodfellow2016deep} that are trained to learn a multi-modal shared latent-space. Such latent-space projection is learned to enable the translation between different modalities, and thus retrieval of TS from different query modalities (Figure~\ref{fig:model_arch} - Training). We can create a dedicated database, indexed using such latent-space projections, which we later use to enable user queries (Figure~\ref{fig:model_arch} - Database). By training encoders to project different input modalities to the same latent space in which the target database TS are stored, we can support diverse query modes (like text, images, sketches) and retrieve matching TS (Figure~\ref{fig:model_arch} - Retrieval).

\subsection{Text-based retrieval}
In our \textbf{text-based retrieval} method, our model allows users to perform TS retrieval via natural language descriptions of desired properties. We will now describe our approach for this.

\subsubsection{\textbf{Architecture}}

To support retrieval functionality across different modes of data, we require an approach to learn consistent representations across modalities. To do this, we implement the CLIP model proposed by \cite{Radford_CLIP} as the deep encoder, and it leverages a set of uni-modal, neural net encoders, pre-trained on large, generalised datasets. The encoders output variable-sized feature representations of their input data, which in this work corresponds to a description of a given TS in the relevant modality. These feature representations are then input into a single layer projection head, which transforms the representations to a consistent dimensionality, as well as allowing fine-tuning through parameter updates during training. Taken together, we can then use a contrastive loss to learn a shared space across the different modalities.

Note that one of the advantages of our model is that for a given entry in the database, the embeddings output by the encoder remains fixed. This means these embeddings can be saved to a look-up table after first evaluation. On retrieval, the single query embedding is calculated and cosine similarities calculated very quickly on-the-fly in a single vectorized operation, which speeds up the retrieval calculation significantly. 

For this text-based retrieval method, we needed to project TS data and natural language text into the same space. We do this by encoding TS data as images, and embedding it in the same space as the text. We will now go over the two encoders that do this.

\paragraph{Image Encoder}
By representing our TS data as images, we can leverage the capabilities of the pre-trained deep convolutional networks (CNN) which have shown remarkable performance on image-based learning tasks. In this work, we use the readily-available ResNet model, introduced in 2015 \cite{He_ResNet}. Its architecture consists of a series of residual blocks, where each block corresponds to a small CNN followed by an element-wise addition with the input (thus learning residual functions), implemented via a skip connection. This aids the stability of the optimisation process - with residual functions being easier to learn than unreferenced functions - and allows a depth of network previously unseen in computer vision. Multiple variants of ResNet exist dependent on their depth; here we use the ResNet50 model, which is made up of 50 layers and strikes a compromise between computational efficiency and model performance. The model was trained on the ImageNet classification dataset \cite{ImageNet}, which consists of 1.28 million training images separated into 1000 classes. It takes as input a 224 $\times$ 224 pixel image, and outputs a corresponding 2048-dimensional vector representation.

\paragraph{Text Encoder}

For the text encoding, we utilise Sentence-BERT (\cite{Reimers_SBERT}) which aims to learn semantically meaningful sentence embeddings, through its sentence transformer models which map collections of words (sentences and paragraphs) to individual embeddings, using a combination of Siamese networks \cite{Bromley_Siamese_Nets} and triplet loss \cite{Schroff_FaceNet}. The model is an alternative to the more widely-known BERT (Bidirectional Encoder Representations from Transformers, \cite{Devlin_BERT}), carrying out tokenization at the sentence-level rather than the original word/sub-word based approach. Through this architecture, the model is able to learn text embeddings in which semantically similar sentences are close together. We use a pre-trained SBERT model, which has been initialised on an English Wikipedia dataset of over 2500M words (through the underlying pre-trained BERT models), and further fine-tuned for meaningful sentence-level embeddings through the SNLI \cite{Bowman_SNLI} and Multi-Genere NLI \cite{Williams_Multi-Genre_NLI} datasets.

\subsubsection{\textbf{Training}}
Next we discuss how the encoders were trained. For a given batch of image, caption pairs and their corresponding embeddings $\bf{v}$ and $\bf{t}$, we first calculate the scaled cosine similarity between all image-text pairs

\begin{equation*}
    \bf{S} \equiv \bf{S}(\bf{t}, \bf{v}^{T}) = \frac{\bf{t} \cdot \bf{v}^{T}}{\tau} \ ,
\end{equation*}

to evaluate the closeness $\bf{S}$ of the respective modalities of data in embedding space. The respective uni-modal similarity for each element is then calculated

\begin{equation*}
    \bf{S}(\bf{t}, \bf{t}^{T}) = \frac{\bf{t} \cdot \bf{t}^{T}}{\tau} \quad\text{and}\quad  \bf{S}(\bf{v}, \bf{v}^{T}) = \frac{\bf{v} \cdot \bf{v}^{T}}{\tau} \ ,
\end{equation*}
which are averaged over to give the target similarity

\begin{equation*}
    \bf{S}_{\text{target}} = \text{softmax} \left( \frac{\bf{S}(\bf{t}, \bf{t}^{T}) + \bf{S}(\bf{v}, \bf{v}^{T})}{2\tau} \right) \ .
\end{equation*}

The final loss $\mathcal{L}$ is calculated averaging over the cross-entropy between each uni-modal embedding and this target similarity

\begin{equation*}
    \mathcal{L} = \frac{\mathcal{L}_{\text{CE}}(\bf{S}, \bf{S}_{\text{target}}) + \mathcal{L}_{\text{CE}}(\bf{S}^{T}, \bf{S}_{\text{target}}^{T})}{2} \ . 
\end{equation*}

which acts to drive the embedded space of each modality to a consistent, shared space. Note the scaling parameter $\tau$ (temperature) used in the above calculation; this is used to scale the similarities such that the sharpness of their predictions can be tuned during validation. Note also that we hold out 10\% of each dataset during training for validation.

\subsection{Sketch-based retrieval}
Now we discuss how encoding was done for sketch-based retrieval. Our sketch-based retrieval enables users to draw a TS, and retrieve a similar one based on the drawn trend and the properties in it (as shown in Figure~\ref{fig:main_figure}.b). While the sketch-based approach can incorporate any statistical property $\xi(\bx)$ in the trend into the search, for simplicity of discussion we focus only on search that incorporates \textit{price volatility} from the sketch, in addition to matching the \textit{price trend}. We use the standard deviation of prices as the measure of volatility. 
We incorporate volatility information into the latent space by computing an \textit{additional} TS associated to the desired property; we compute the volatility in a fixed window of the input TS, and slide the window across the original TS to generate a new ``volatility TS". By doing this we can use the same encoding architecture used on the original TS. This idea of displaying information in derived TS is seen in analytics for finance with properties like moving averages, and Relative Strength Index (RSI). We apply use it to encode and store such information. In this way we can preserve it explicitly into the latent space for informing subsequent database retrieval. This gives us a general way of storing other such properties($\xi(\bx)$) that can be computed from the TS.

\subsubsection{\textbf{Training}}
All input historical TS $\bx$ are normalized to be between $[0,1]$ using a Min-max scaler. Additionally, given a TS $\bx$, we define its \textit{price volatility} in terms of a new TS  $\sx \in [0, 1]^{[L]}$ where each data point $j \in \sx_i$ is the volatility of the neighboring $2*m$ points ( we use $m=4$):
$
 \sx_i = \sigma(\bx_{i-m:m+i})
$ We will refer to the \textit{price trend}(original) TS as simply ``TS'', and the \textit{price volatility} TS computed from it as volatility-TS.

For searching based on the trend and the volatility, we train two identical (and separate) fully-connected auto-encoders for the TS and volatility-TS. To incorporate the trend information, we train a classic fully-connected autoencoder(AE) $E_t(\bx)$ that encodes the TS $\bx$ into the latent-space with the following reconstruction loss: 
$$
    \min_{i \in \mathbb{D}} |\bx_i - \bar{\bx}_i |^{2}_{2}
$$
where $\bar{\bx}$ is the reconstructed TS. To include \textbf{volatility} information into the search, we train another deep AE $E_v(\sx)$ to encode $\sx$ into the latent-space, with a similar reconstruction loss: 
$$
    \min_{i \in \mathbb{D}} |\sx_i - \bar{\sx}_i |^{2}_{2}
$$
where $\bar{\sx}$ is the reconstructed volatility TS. 

The overall encoder $E$ for the sketch-based retrieval simply concatenates and normalizes the embeddings from $E_t(\bx)$ and $E_v(\sx)$. 
With respect to the network structure of the AE, the input size is 30, and there are 3 hidden layers in the encoder of size 512, 256, and 16; the last layer is the latent space of size 16. We normalize the latent space before decoding in the AE using the same structure as the encoder albeit in reverse. All layers include trainable bias weights, and use ReLU (Rectified Linear Unit) as the non-linearity function, except for the last layer of the decoder which does not include ReLU computation.

\subsubsection{\textbf{Database}} In the database construction we compute the AE output (latent space) for each TS $\bx_i \in \mathbb{D}$. In particular, for each TS we can compute its embedding $e$, being the concatenation of $E_t(\bx_i)$ and $E_v(\sx_i)$, and we store it into a database structure. 
The database maintains the embedding and original TS $<\boldsymbol{e}, \bx>$ to provide a subsequent efficient search. In particular, we use Facebook AI Similarity Search (FAISS) ~\cite{FAISS}, a library for very fast searching over vector spaces, for indexing and lookup. 

\subsubsection{\textbf{Retrieval}}
To retrieve TS using \textit{trend} and \textit{volatility} information in the input TS, we use the AE to generate a query vector to search the FAISS index. Given a user sketch that contains a trend $\bx$, we compute the volatility TS $\sx$, and then compute the embedding $\bar{\boldsymbol{e}}$ for the query using the encoders. We can then use this embedding to efficiently retrieve a subset of TS $\bx_i \in \mathbb{D}$ whose cosine similarities are the highest in the database.

In the experimental section we show how these encoders can preserve information for both \textbf{trend} and \textbf{volatility}. We also show how this approach is much faster than existing baseline approaches for comparable TS retrieval using the same modality (TS sketches). Most importantly, while existing consumer TS databases require the user to specify a rigid definition for matching properties such as a trend approximation (e.g., second-degree polynomial) we found that our AEs naturally approximate and match any trend in the input sketch TS. This provides not only a faster retrieval mode, but also more flexibility to the end user.

\section{Empirical Analysis}

\subsection{Evaluation Measures}

\subsubsection{Text-Based Retrieval Measures}
We use three metrics to determine performance, with evaluation carried out on the validation set of unseen TS images for both in-sample (used during training) and out-of-sample queries. Each query is a single statement about a particular feature, with a roughly equal number of queries for each regime. We evaluate retrieval accuracy using Rank@9. This is recorded for both in-sample and out-of-sample settings. In addition, we define a diversity metric, which measures the number of distinct TS images returned across all queries relative to the total number of images returned. Thus a higher measure of diversity implies a greater fraction of images in the database are able to be retrieved through prompt variation. For this metric we record the aggregated value across both querying settings.

\subsubsection{Sketch Based Retrieval Measures}
To evaluate the sketch-based retrieval method, we automatically generate TS queries (as stand-ins for sketch TS) by using 302 TS queries from the test-set (not given during training to the AE). To these, we also added noise and a circular shift to the right by 5 steps to generate TS that are less similar to training data, and query for best matches. For evaluating the retrieval performance of each method, we consider measures that match the query sketch with the returned TS. The measures we compute to compare the performance between baselines and AE are as follows:
\textbf{(1) MAPE}: Is the Mean Absolute Percentage Error(MAPE) between the trend and retrieved TS. This computes the average ratio between point-wise error and the true value. Note that we display it as a ratio, and not as a percentage. For each method, we average this value over all the test queries (all top-k results returned per query).
\textbf{(2) CORR}: Is the Pearson's correlation coefficient which is between $[-1,1]$, and measured between the query TS and retrieved TS. We average the values over all queries. 
\textbf{(3) Computational Time} (mean and standard deviation): This is computed over all  queries. This time includes the time taken to compute the embedding of each query if the method requires an embedding.

\subsection{Baselines}

\subsubsection{Text-Based Retrieval Baselines}\label{text_retrieval_baselines}

Given the limited number of previous works in this area, no benchmarks have yet been established. A key contribution of this work is therefore to provide an appropriate benchmark for comparison in future studies. Given the lack of such approaches here, we implement two simple baselines which leverage traditional NLP approaches for calculating embeddings with text.
\\
(1) Neural Net Classifier (Classifier): Here we re-formulate the retrieval problem as a classification problem. We first compute 100-dimensional embeddings for all the words in our vocabulary, through the use of a traditional NLP encoding algorithm - Word2Vec \cite{Mikolov_Word2Vec} - which is pre-trained on the Text8 dataset of 100 million bytes of plain text taken from Wikipedia. Ground truth image captions are then converted into numeric vector representations through averaging over the respective word embeddings, from which cosine similarities are calculated for each text-based query embedding in the training set. Given an input query vector, the network outputs the predicted cosine similarities for each sample in the database. We optimise the network with respect to the MSE loss during training. The output similarities can then be sorted into descending order and retrieval carried out. Since the network architecture is fixed such that each output node corresponds to a specific sample in the database only in-sample retrieval is applicable; we therefore carry out both training and retrieval on the validation set.
\\
(2) Word2Vec-UMAP: As an alternative baseline, we replace our SOTA deep learning encoders with traditional approaches to representation learning. The respective uni-modal encoders are replaced with the following: Text encoder $\rightarrow$ Word2Vec and Image encoder $\rightarrow$ UMAP. 
\\
As with our encoder-based model, we align the embedding using the contrastive loss and a single layer MLP projection head for each modality. The final aligned space has a dimensionality of 64, with the text and image embeddings being 100 and 2048-dimensional respectively.

\begin{table}[t]
    \centering
    \begin{tabular}{L{0.15\textwidth}T{0.08\textwidth}T{0.12\textwidth}T{0.08\textwidth}}    
    \toprule
        Method & Rank@9 In-sample & Rank@9 Out-sample & Diversity \\ 
        \hline
        Classifier & 0.59 & 0.43 & 0.40 \\
        Word2Vec-UMAP & 0.37 & 0.35 & 0.05 \\
        Ours (no filtering) & 0.65 & 0.51 & 0.18 \\ 
        Ours (filtered) & 0.92 & 0.71 & 0.50 \\
        Ours (filtered+) & \bf 0.96 &  \bf 0.89 & 0.30 \\ 
        \bottomrule
    \end{tabular}
    \caption{\small Retrieval results on the synthetic dataset for the following methods: neural classifier, outputting retrieval probabilities for all TS in a database given a text embedding; contrastive learning with traditional representation learning approaches (UMAP and Word2Vec); Deep Encoding Networks with three filtering variants. Filtered+ denotes a post-process filtering step alongside GPT augmentation of captions.}
\label{tab:synthetic_text_retrieval}
    \vspace{-0.5in}
\end{table}

\subsubsection{Sketch Based Retrieval Baselines}
We evaluate our sketch-based retrieval against 3 baseline methods which are as follows :\\
(1) Brute Force Search (BF): this method takes a query TS, and compares it against all entries in the dataset. It chooses the top-k best matches by Euclidean distance between the query and each TS of the database; this is the L2-norm of the error. We ignore the volatility information for this baseline.\\
(2) Brute Force Search over TS and Volatility-TS (BF\_avg): In this method, we repeat the brute force search, except we compute Euclidean distance of the error for both the original TS and the volatility-TS, and average the distances. This averaged distance is used to find the top-k closest matches from the database.\\
(3) UMAP: We compute UMAP embeddings for TS and Volatility-TS for the dataset and store it in the FAISS index for lookup by concatenating and normalizing the vectors of both embeddings. The dimensionality of the embedding space matches that of the AE (16). After indexing, each query is converted into the umap embedding of it's TS and volatility-TS and that is used to search in the FAISS index. In this work, all TS are of the same length, so we do not use dynamic time warping (DTW), as it reduces to just euclidean distance for same-length TS, which is what the BF methods use.

\subsection{Results}
We present results of our two query modalities (natural language, and sketch TS) and compare them with results from baselines.

\subsubsection{Text Based Retrieval Results}

\paragraph{Synthetic Data}

The retrieval results on the synthetic dataset are shown in Table \ref{tab:synthetic_text_retrieval}. We evaluate three variations of our model against the text-based retrieval baselines (Classifier, Word2Vec-UMAP); all three variations of our Deep Encoding Network TS retrieval approach outperform the baselines. In addition, both baseline models are considerably less flexible than our proposed method. For the classifier network, this lack of flexibility is due to the requirement for a fixed network architecture, with each output node corresponding to a single entry in the retrieval database. As such, any new entries in the database can only be incorporated through complete re-training with the augmented data. On the other hand, the Word2Vec-UMAP baseline is limited not by any structural constraints but rather due to the computational expense of the UMAP approach to representation learning in a very high-dimensional space. As such, we could only train the embedding on 4000 images, with each flattened image being represented by a 50,176-dimensional vector. We varied each of the nearest neighbours, minimum distance and number of components parameters of the UMAP model, but all had little impact on final performance.

\begin{table}[t]
    \centering
    \begin{tabular}{L{0.15\textwidth}T{0.08\textwidth}T{0.08\textwidth}T{0.12\textwidth}}
    \toprule
        Method & Dataset Size & Rank@9 In-sample & Rank@9 Out-sample \\ 
        \hline
        Unfiltered & 1500 & 0.56 & 0.59  \\ 
        Unfiltered-Retrained & 1500 & 0.69 & 0.60  \\ 
        Filtered & 1500 & 0.59 & 0.51  \\ 
        Filtered-Retrained & 1500 & 0.48 & 0.48\\ 
        Unfiltered-Retrained & 3000 & 0.75 & 0.62 \\ 
        Filtered-Retrained & 3000 & 0.55 & 0.51 \\ 
        \bottomrule
    \end{tabular}
    \caption{\small Retrieval results on the historic dataset with our Deep Encoding Networks. Note Filtered/Unfiltered here refers to GPT-augmented filtering of trend, volatility and shock probability.}
    \label{tab:historic_text_retrieval}
\end{table}

\paragraph{Historical Data}

Table~\ref{tab:historic_text_retrieval} shows the results for the historical dataset. These are evaluated for both the pre-trained model from the synthetic data study, and when re-trained on the new dataset.  Each stock is associated with a volatility regime determined by the liquidity \cite{bouchaud2018trades}. For comparison, an alternative dataset is constructed in which volatility ground truth is determined by the filtering approach discussed in section \ref{database_construction}, rather than liquidity. We find that carrying out volatility filtering on the dataset prior to evaluation is not essential, and often hinders accuracy, even when evaluating on the original model trained on the synthetic dataset. Whilst the performance is competitive for the pre-trained model, re-training on the new data is found to yield better results, particularly as the dataset size is increased. Given the performance improvement when increasing the dataset size, we see no reason to think the model will not perform comparably on historic stock data to that seen in the synthetic case as the dataset is scaled.

\begin{table*}[t]
\centering
\begin{tabular}{lllllll}
\toprule
noise,shift,k & measure & BF & BF\_avg & UMAP & AE \\
\midrule
 & TS-(MAPE,CORR) & (\textbf{8.41E-02, 9.83E-01}) & (\textbf{8.41E-02, 9.83E-01}) & (2.28E-01, 8.43E-01) & (9.98E-02, 9.71E-01)\\
0.05, 0, 1 & Vol-(MAPE,CORR) & (2.20E-01, 8.02E-01) & (2.20E-01, 8.02E-01) & (2.17E-01, 7.34E-01) & (\textbf{2.09E-01, 8.04E-01}) \\
 & Time-(mean,std-dev) & (4.27E-02, 1.48E-02) & (5.08E-02, 1.61E-02) & (7.71E+00, 6.60E-01) & \textbf{(1.90E-02, 8.16E-03)} \\  \hline
 &TS-(MAPE,CORR) & (\textbf{2.11E-01, 8.96E-01}) & (2.17E-01, 8.95E-01) & (3.61E-01, 7.49E-01) & (2.76E-01, 8.24E-01) \\
0.05, 5, 1 & Vol-(MAPE,CORR) & (3.77E-01, 5.73E-01) & (3.37E-01, \textbf{5.97E-01}) & (3.30E-01, 5.30E-01) & (\textbf{3.20E-01}, 5.44E-01) \\ 
 & Time-(mean,std-dev) & (1.97E-01, 1.33E-02) & (2.32E-01, 1.35E-02) & (7.82E+00, 8.29E-01) & \textbf{(1.87E-02, 7.66E-03)} \\ \hline
 &TS-(MAPE,CORR) & (\textbf{1.94E-01}, 8.92E-01) & (1.99E-01, \textbf{8.86E-01}) & (3.07E-01, 7.46E-01) & (2.33E-01, 8.33E-01) \\
0.05, 0, 3 & Vol-(MAPE,CORR) & (3.57E-01, 5.37E-01) & (3.00E-01, 6.00E-01) & (\textbf{2.64E-0}1, 6.24E-01) & (2.66E-01, \textbf{6.25E-01}) \\ 
 & Time-(mean,std-dev) & (1.97E-01, 1.38E-02) & (2.34E-01, 1.55E-02) & (7.64E+00, 1.13E+00) & \textbf{(1.81E-02, 8.52E-03)} \\ \hline
 &TS-(MAPE,CORR) & (\textbf{2.60E-01, 8.53E-01}) & (2.66E-01, 8.50E-01) & (3.79E-01, 7.18E-01) & (3.04E-01, 7.93E-01) \\
0.05, 5, 3 & Vol-(MAPE,CORR) & (4.21E-01, 4.64E-01) & (3.78E-01, 5.00E-01) & (\textbf{3.36E-01, 5.37E-01}) & (3.41E-01, 5.16E-01) \\
 & Time-(mean,std-dev) & (1.87E-01, 8.90E-03) & (2.22E-01, 9.51E-03) & (7.66E+00, 1.19E+00) & \textbf{(1.86E-02, 8.80E-03)} \\
\bottomrule
\end{tabular}
\caption{\small Average of a set of TS matching measures over test-set queries with Gaussian noise ($\mathcal{N}(\mu = 0, \sigma = 0.05)$) and circular shifts for the baseline methods (BF, BF\_avg,UMAP) and AE method.}
\label{tab:sketch_retrieval_results}
\end{table*}

\subsubsection{Sketch Based Retrieval Results}
With respect to our sketch-based retrieval approach, we present the results of the baseline methods (BF,BF-avg, UMAP) and the Auto-Encoder (AE) method in Table \ref{tab:sketch_retrieval_results}. For the measure MAPE, lower is better, while for CORR (correlation) higher is better. The first column describes the test scenario defined by the noise added, the circular-shift steps (shift) which is the number of steps by which the TS is rotated, and the number of best matches returned (k).

From the data, we can see that the best fit in terms of matching the TS-trend data (``TS-(MAPE,CORR)'' measures) comes from the BF method, which is no surprise since it painstakingly computes the Euclidean distance between query TS and every TS in the database before taking the least distant (least error) one. However, this distance between TS does not translate to matching the volatility information, and this is seen in the results; the other methods which consider volatility during search do better in matching the volatility-TS measures (``Vol-(MAPE,CORR)'').
Our AE typically does better than the baseline methods in matching the volatility TS ( for MAPE and CORR measures), and when it is not the best, it is very close to the best value by UMAP. Importantly, the AE method takes significantly less time than all the other methods. The UMAP process takes the most time, despite UMAP and AE both using FAISS with the same sized latent space. This time difference between UMAP and AE is due to the UMAP encoding time; the time to project a new query into the same latent space as the database.

\section{Conclusions}
In summary, our paper presented a framework for multi-modal storage and retrieval of financial TS data such that pertinent information for finance is preserved. In particular, we developed user-friendly query interfaces that accept natural language text or TS sketches for retrieval. Our methods use deep encoder networks to map the multi-modal data into a lower-dimensional latent space, and preserve essential TS properties. Experimental results support the approach's computational efficiency and retrieval accuracy. Thus we demonstrated the utility of latent-space projections for retrieval of financial time-series data while supporting more intuitive query modalities.

\section*{Disclaimer}
This paper was prepared for informational purposes by
the Artificial Intelligence Research group of JPMorgan Chase \& Co\. and its affiliates (``JP Morgan''),
and is not a product of the Research Department of JP Morgan.
JP Morgan makes no representation and warranty whatsoever and disclaims all liability,
for the completeness, accuracy or reliability of the information contained herein.
This document is not intended as investment research or investment advice, or a recommendation,
offer or solicitation for the purchase or sale of any security, financial instrument, financial product or service,
or to be used in any way for evaluating the merits of participating in any transaction,
and shall not constitute a solicitation under any jurisdiction or to any person,
if such solicitation under such jurisdiction or to such person would be unlawful.

\balance
\bibliographystyle{ACM-Reference-Format}
\bibliography{refs}

\end{document}